\documentclass[final,5p,times,twocolumn]{elsarticle}

\usepackage{hyperref}

\journal{Computers \& Security}

\usepackage[usenames,dvipsnames]{color}

\usepackage[utf8]{inputenc}

\usepackage{graphicx}
\usepackage{amsmath}
\usepackage{amsfonts}
\usepackage{amssymb}
\usepackage{array}
\usepackage{multirow}
\usepackage{epstopdf}
\usepackage{caption}
\usepackage{subcaption}
\usepackage{xcolor}

\usepackage{url}
\usepackage{afterpage}
\usepackage{booktabs}

\usepackage{changepage}

\usepackage{url}
\makeatletter
\g@addto@macro{\UrlBreaks}{\UrlOrds}
\makeatother

\begin{document}

\begin{frontmatter}

\title{Reversing the Irreversible:\\A Survey on Inverse Biometrics}


\author[1]{Marta Gomez-Barrero\corref{cor1}}
\ead{marta.gomez-barrero@h-da.de}
\author[2]{Javier Galbally}
\ead{javier.galbally@ec.europa.eu}

\cortext[cor1]{Corresponding author}

\address[1]{da/sec - Biometrics and Internet Security Research Group, Hochschule Darmstadt, Germany}
\address[2]{European Commission - DG-Joint Research Centre, E.3, Italy}

\begin{abstract}

With the widespread use of biometric recognition, several issues related to the privacy and security provided by this technology have been recently raised and analysed. As a result, the early common belief among the biometrics community of templates irreversibility has been proven wrong. It is now an accepted fact that it is possible to reconstruct from an unprotected template a synthetic sample that matches the bona fide one. This reverse engineering process, commonly referred to as \textit{inverse biometrics}, constitutes a severe threat for biometric systems from two different angles: on the one hand, sensitive personal data (i.e., biometric data) can be derived from compromised unprotected templates; on the other hand, other powerful attacks can be launched building upon these reconstructed samples. Given its important implications, biometric stakeholders have produced over the last fifteen years numerous works analysing the different aspects related to inverse biometrics: development of reconstruction algorithms for different characteristics; proposal of methodologies to assess the vulnerabilities of biometric systems to the aforementioned algorithms; development of countermeasures to reduce the possible effects of attacks. The present article is an effort to condense all this information in one comprehensive review of: the problem itself, the evaluation of the problem, and the mitigation of the problem. The present article is an effort to condense all this information in one comprehensive review of: the problem itself, the evaluation of the problem, and the mitigation of the problem. 


\end{abstract}


\end{frontmatter}


\section{Introduction}
\label{sec:intro}


\begin{adjustwidth}{0.5cm}{0.5cm}
``\emph{Nothing in this world can be said to be irreversible, except
death and taxes.}'' - Benjamin Franklin\footnote{Paraphrase of a
quote usually attributed to Benjamin Franklin who, in 1789, wrote a
letter stating: ``\emph{Our new Constitution is now established, and
has an appearance that promises permanency; but in this world
nothing can be said to be certain, except death and taxes}''}.
\end{adjustwidth}

\vspace{0.15cm}

Since the first works on biometric verification, increasingly accurate and
time efficient biometric recognition systems have been proposed over
the years.
These works have allowed a wider deployment of biometric
authentication techniques, including border control
\cite{labati16biometricsBorderControlSurvey}, smartphone
authentication \cite{patel16continuousAuthentication}, mobile
payments \cite{meng15surveyBiometricsMobile}, or law enforcement and
forensics \cite{jain15biometricsForensics}. With this generalised
use of biometric systems, new concerns have arisen regarding their
potential weaknesses. Among these vulnerabilities, one stands out
due to the serious risks that it poses: the possibility to use a
compromised biometric template to recover the raw bona fide sample
from which it was generated.

A classical biometric system acquires a probe sample of the biometric
characteristic of an individual, extracts salient features from the
sample (i.e., biometric template), and compares the extracted
features against the previously enrolled reference template in order to \textit{verify} a
claimed identity or to \textit{identify} an individual. For security
and privacy reasons, biometric systems typically do not store the raw biometric data, which may disclose sensitive
information about the subjects (e.g., race, gender, diseases, etc.)
Rather, they store the extracted template containing the most
discriminative information about the individual, relevant for
recognition purposes. Such a protection approach involves one major
assumption: biometric templates do not contain enough information in
order to be reversed engineered and to recover from them the bona fide
sample.

The key question is: \textit{Are biometric templates really
irreversible?} Until not long ago, the answer to that question was:
yes, biometric templates are irreversible. It was a common belief that the features extracted from
biometric samples did not reveal enough information about the
underlying biometric characteristic and its owner in order to be
exploited with malicious purposes \cite{IBG02imagesFromTemplates}.
However, in 2001, Hill put forward the problem of storing biometric
templates without the right protection measures
\cite{hill01AttackDB}. This was also the first work to consider the
possibility of reconstructing the bona fide samples given only the
information stored in the template. Shortly afterwards, in 2003,
Bromba formally studied the issue of template reversibility
\cite{bromba03invBio}. In particular, he explored whether
``biometric raw data can be or not reconstructed from template
data''. In other words, he challenged the established knowledge by
analysing whether it was possible to \textit{reverse the
irreversible}. In that article, he reached three main conclusions:

\begin{table*}[t]
\begin{small}
\begin{center}

\caption{Summary of synthetic biometric sample generation methods
with the main applications they are used for. References are not an
exhaustive list, just indicative examples of works where a given
type of methods has been used for a specific
application.}\label{tab:apps} \centering
\begin{tabular}{lccccc}
\toprule
\multirow{2}{*}{} & \textbf{Transformation} & \textbf{Combination} & \textbf{Morphing} & \textbf{Generation} & \textbf{Inversion}  \\
 &  \textbf{methods} & \textbf{methods}& \textbf{methods}& \textbf{methods} & \textbf{methods}\\ \midrule
Enrolment data  & \cite{galbally14OnOffSign} & \cite{vanDam15face3Drec} & - & - & - \\
Training data  & \cite{masi16faceDuplicated} & - & \cite{banerjee17faceCombination} & \cite{Mai-DeepFaceReconstruction-PAMI-2018} & - \\
Synthetic benchmarks  & - & - & - & \cite{maltoni2004generation,papi16alteredFingSynth} & - \\
System testing  & - & \cite{cappelli2007synthFpeval,marques2000effects} & - & - & -\\
Pseudo identities  & - &- & \cite{ross13mixingFingerprints,li13combiningFingerprints} & - & - \\
Vulnerability studies  & \cite{popel07signatureSynth} & \cite{scherhag16PADunitSelection}  & \cite{ferrara2014magicPass,scherhag17morphingScanned} & \cite{papi16alteredFingSynth,yadav2019synthIrisRaSGAN} & \cite{Galbally2013,Mai-DeepFaceReconstruction-PAMI-2018,kim2018reconstructionFpGAN} \\
\bottomrule
\end{tabular}\vspace{-0.5cm}
\end{center}
\end{small}
\end{table*}

\begin{itemize}
\item There are cases where raw data are very similar to template
data by definition, and therefore can hardly be distinguished.

\item Often the reconstruction is possible to a degree which is
sufficient for misuse (i.e., the reconstructed sample is accepted by
the biometric system).

\item Even if reconstruction should not be possible in specific
cases, misuse of templates remains possible.
\end{itemize}

Following those pioneering works, several studies have supported the
same findings, arising serious concerns regarding the soundness of
the aforementioned irreversibility assumption for different
characteristics, including fingerprints
\cite{Cappelli07PAMIReconstruction}, face
\cite{adler04quantizedFaces}, or iris
\cite{venugopalan11irisImageFromBinary}. Even the latest deep
learning approaches have been shown to be vulnerable to these
attacks
\cite{Akhtar-DeepAttacksSurvey-Access-2018,Mai-DeepFaceReconstruction-PAMI-2018}.
In all those works, the information stored in reference templates is
exploited in order to generate synthetic samples in so-called \emph{reversibility attacks}. These samples can be
subsequently used to: 1) launch masquerade attacks (i.e.,
impersonating a subject), thereby decreasing the security of the
system; or 2) to derive information from its owner, thereby
threatening the subject's privacy.


In order to deal with those concerns, the ISO/IEC standard 24745 on biometric information protection \cite{ISO-IEC-24745-2011} specifies irreversibility as one of the major requirements for templates to be used within biometric systems in order to grant the privacy protection data subjects are entitled to in the new EU General Data Protection Regulation \cite{EU-Regulation-DataPrivacy-160427}. Recently, this official recognition of the key importance of the irreversibility of templates in biometrics has fostered and strengthened even more the interest that the biometric community has shown on the study of this field over the last 15 years. This interest has led to: 1) new reconstruction
algorithms for different biometric characteristics
\cite{ross07RecFingerprints,mohanty07InvBiomFace};
2) evaluation methodologies to determine the risk posed by these
reconstruction approaches \cite{Galbally2013,marta13handInvJournal}; and
3) new countermeasures to protect the systems against this potential
threat
\cite{Rathgeb11e}.
All these works have constituted a new research area commonly
referred to as \textit{inverse biometrics}.

Nowadays, inverse biometrics has become a well-established
research field with a large number of publications in journals,
conferences and media, that require a significant condensation
effort to form a clear picture of the state-of-the-art. The present
paper represents the first survey carried out in this active area to
review the progress achieved, presenting in a comprehensive manner
the different inverse biometric methods proposed so far. The article
attempts to be not just a simple enumeration of papers, but to
categorise algorithms according to objective parameters, also
discussing the security and privacy implications of each of them. 

In brief, the paper is thought as a tool to provide biometric
researchers, either newcomers or experts in security related aspects
of this technology, an overall picture of the current panorama in
inverse biometrics. It also aims at discussing the very unique
security and privacy threats posed by these algorithms and how they
can be evaluated and mitigated. Although the work is thought to be
self-contained, some previous general knowledge on biometrics can
help to better understand several of the concepts introduced in the
article.

\section{Synthetic Biometric Samples Generation}
\label{sec:synthSamp}

\begin{figure*}[t]
\centering
\begin{minipage}[c]{0.95\linewidth}
  \centering
  \centerline{\includegraphics[width=.99\linewidth]{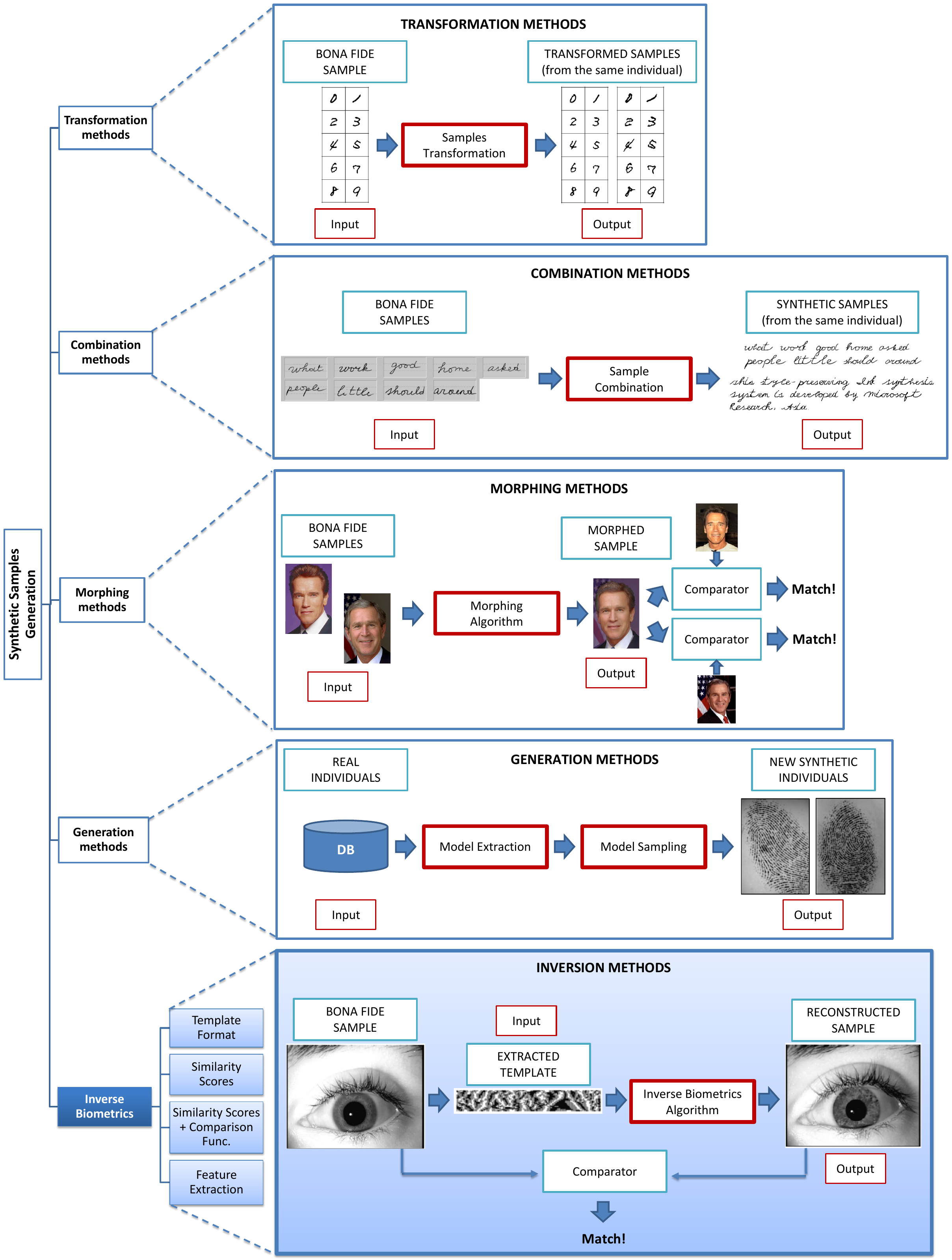}}
\end{minipage}
\caption{Classification of the methods for synthetic biometric
samples generation. The methods that are the main focus of the
present review (i.e., inverse biometrics) are highlighted in blue
and classified according to the knowledge required to be carried
out. Images have been taken from
\cite{mori00handwriting,lin07SyntheticHW,cappelli2004sfinge,Galbally2013}.}
\vspace{-0.5cm} \label{fig:SynthBioClassif}
\end{figure*}

As already introduced, this article is focused on the review of
inverse biometric methods and their security and privacy related
issues. In order to put inverse biometrics into context, the present
section gives a general overview of the broader field of synthetic
biometric samples generation. As a graphical aid,
Fig.~\ref{fig:SynthBioClassif} shows the overall taxonomy considered
in the article regarding the different techniques proposed so far to
produce synthetic samples, together with their main applications in
Table~\ref{tab:apps}.



Historically, the \textit{manual} production of \emph{physical}
biometric characteristics such as fingerprints, signatures, or forged
handwriting has been a point of concern for experts in the
biometric field from a forensic point of view
\cite{wehde1924fingerForge,osborn1946questionedDocs}. More recently,
such physically produced synthetic characteristics have been largely
utilised for vulnerability and presentation attack (a.k.a. spoofing)
assessment studies in characteristics such as the fingerprint
\cite{galbally09FPsPRLda}, the iris \cite{cui04synthIrises}, or the
face \cite{adler03hillfaces}. 

\afterpage{\clearpage}

However, it was not until the digital revolution that lead to the
large development of biometric recognition technology in the 90's,
when the automatic generation of \textit{digital} synthetic samples
started to be widely studied \cite{cappelli2004sfinge,blanz993DfaceSynth,lefohn03ocularist,yanushkevich05inverseBio}. This field has observed a big
progress in the very recent past thanks to the appearance of deep
learning generative methods that are able to produce novel samples
from high-dimensional data distributions, such as images. This is
the case for instance of the popular Generative Adversarial Networks
(GANs) \cite{Goodfellow-GANs-NIPS-2014}, which have shown great
potential to produce face images  \cite{karras17GANface} and 3D models \cite{gecer2019ganfit}, iris images \cite{kohli17GANiris,yadav2019synthIrisRaSGAN}, retina images
\cite{kaplan17GANretina}, and fingerprints \cite{bontrager2017deepmasterprint,kim2018reconstructionFpGAN}, or of the
autoregressive models that have been used to produce synthetic
speech \cite{oord16GANaudio} and highly realistic cursive
handwriting \cite{graves13autoregressiveHW}.

\begin{figure*}[t]
\centering
\begin{minipage}[b]{0.7\linewidth}
  \centering
  \centerline{\includegraphics[width=.95\linewidth]{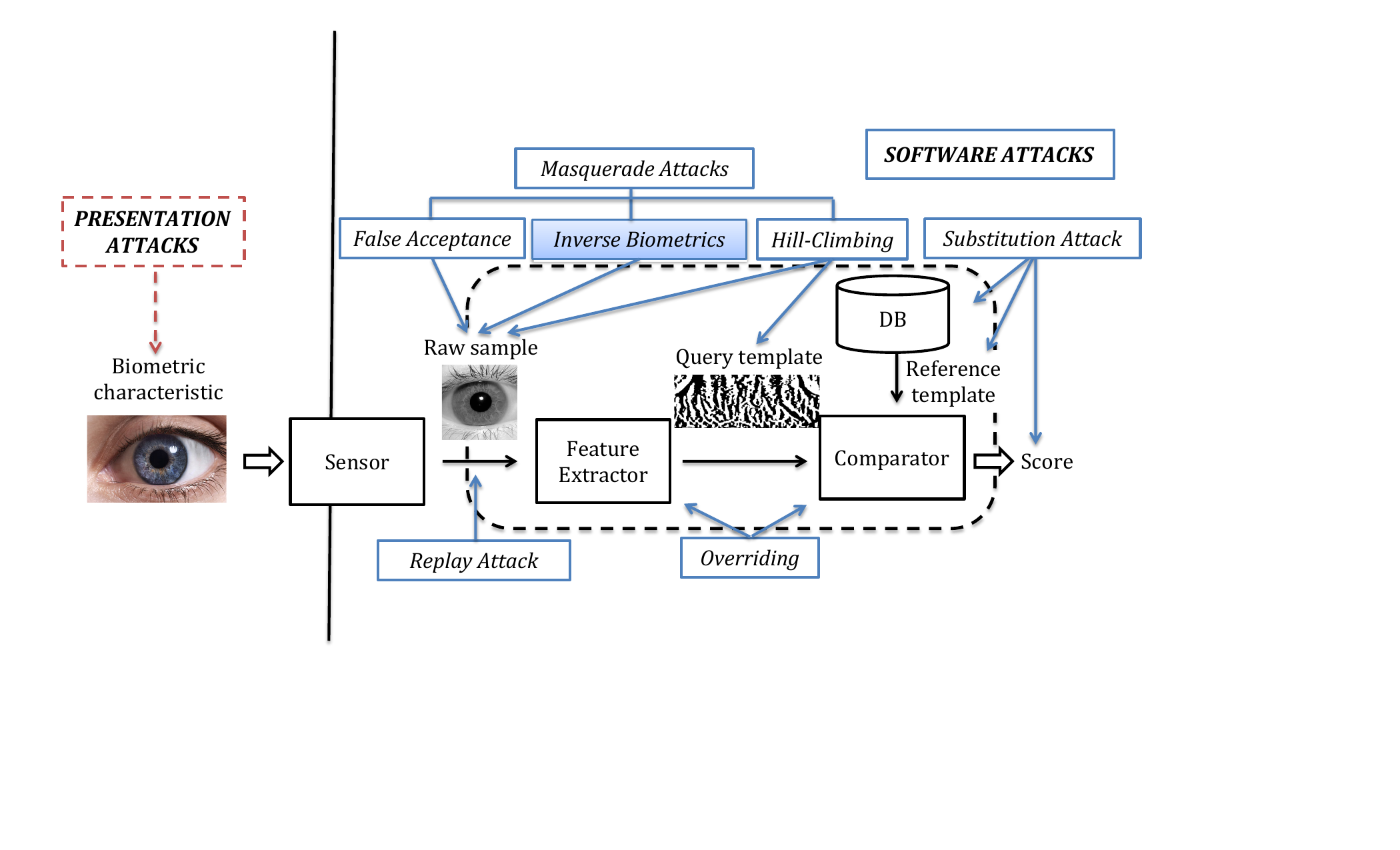}}
\end{minipage}
\caption{General classification of attacks on biometric recognition systems, which can be broadly divided into presentation and software attacks.} \label{fig:bioattacks}
\end{figure*}

\begin{figure*}[t]
\centering
\begin{minipage}[b]{0.8\linewidth}
  \centering
  \centerline{\includegraphics[width=.99\linewidth]{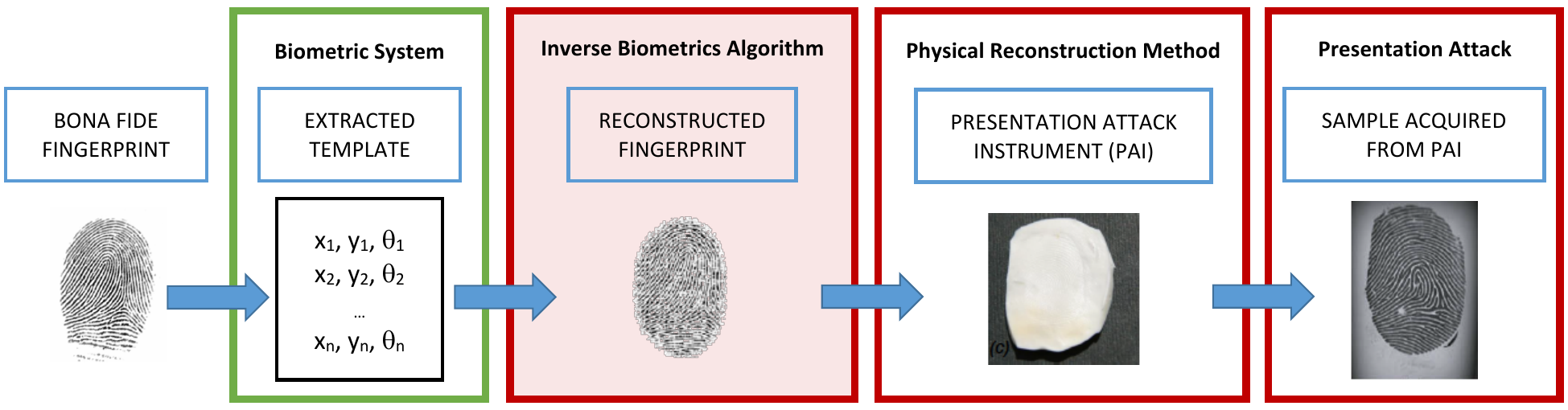}}
\end{minipage}
\caption{Example of how a compromised template is used to
reconstruct, through an inversion algorithm, a biometric
sample which can lead to other type of stronger threats such as
presentation attacks. Images extracted from
\cite{galbally09FPsPRLda,Galbally2008_ICPR}.} \vspace{-0.5cm}
\label{fig:gummyFingers}
\end{figure*}

The main reason behind the significant research efforts dedicated in recent times to the generation of synthetic biometric samples are the numerous applications of this field, especially in the context of vulnerability studies.
As mentioned in the introduction, probably the widest use given to synthetic biometric samples is the security assessment of systems, within the more general framework of adversarial machine learning \cite{biggio15advBioRecognitionSurvey}. For completeness, we present in Fig.~\ref{fig:bioattacks} the most common vulnerabilities classification considered in biometric literature \cite{ratha01securityPrivacy}. As can be seen, biometric attacks may be widely divided into: 1) presentation attacks, which are performed against the biometric sensor using some type of physical artefact (e.g., a gummy finger in the case of fingerprint-based systems); 2) software attacks, which are directed to some of the internal components of the system or the communication channels between them. To be effective, software attacks need 1) some level of knowledge about the internal functioning of the system, and 2) access to some of the internal components of the system. Presentation attacks, on the other hand, are carried out in the physical world and do not require any knowledge about the system or access to restricted components, only to the sensor. As such, presentation attacks pose in general a greater threat to biometric systems. As will be discussed later, one of the big challenges posed by inversion algorithms is that they have the potential to transform software attacks into presentation attacks (see Fig.~\ref{fig:gummyFingers}). Some examples of studies that have used synthetic biometric data to attack biometric system include the use of artificial samples to maximise the similarity score of a particular recognition system \cite{alegre2012syntheticSpeech}. Also, synthetically reconstructed images which would be positively matched to the stored reference, can be submitted to impersonate the enrolled subjects \cite{Galbally2013}. And more recently, it was shown that synthetic iris images produced by GANs were able to fool even state-of-the-art presentation attack detection methods
\cite{kohli17GANiris,yadav2019synthIrisRaSGAN}.

In addition to vulnerability assessment works, synthetic biometric samples have also been applied to: 1) create enrolment data: especially for behavioural characteristics with large intra-class variability where the scarcity of enrolment data is a key limiting factor to obtain low recognition error rates \cite{galbally14OnOffSign}; 2) create training data: to complement the training of some of the current most popular techniques such as Deep Neural Networks (DNNs) \cite{masi16faceDuplicated,banerjee17faceCombination}; 3) create synthetic benchmarks: not subjected to the framework of personal data protection that can therefore be shared among researchers to estimate the accuracy of systems under a common benchmark \cite{sumi06synthFaceDB}; 4) system testing: the impact of some specific issues on system performance can be investigated thanks to the fine parametric control provided by synthetic samples \cite{cappelli2007synthFpeval,sumi06synthFaceDB}; 5) pseudo-identities: in order to preserve the privacy of the subject, enrolled templates could be substituted by templates generated from synthetic samples, which discard non-discriminate
private information such as the gender \cite{ross13mixingFingerprints}; 6) entropy studies: synthetic samples generated from models can help to determine the individuality of a particular biometric modality \cite{hollingsworth09bitsIris,pankanti02fpIndiv}.

These and other applications have fostered over the last years a
growing interest towards the development of new methodologies to
synthesise biometric samples for different characteristics
\cite{bioSynthesisbook06}, such as fingerprints
\cite{Cappelli07PAMIReconstruction},
face \cite{sumi06synthFaceDB}, iris
\cite{Galbally2013},
voice
\cite{schroeter08tts},
handwriting \cite{lin07SyntheticHW}, speech mouth dynamics
\cite{yamamoto1998lipSynthesis}, signature
\cite{galbally12PRssiggeI},
mouse dynamics \cite{nazar08invBioMouse},
or keystroke dynamics \cite{rashid09invBioKeystroke}.

From a general perspective, methods to produce synthetic biometric
samples can be broadly divided into five categories, depending on:
1) the input to the method; 2) the approach followed; and
3) the type of synthetic data created. Those five types,
depicted in Fig.~\ref{fig:SynthBioClassif}, will be described in the
following. 

\textbf{Transformation methods}: starting from one or more bona fide
samples of a given subject, and applying different transformations,
these methods produce different synthetic (or transformed) samples,
which belong to the same subject. Different approaches have been
proposed for face
\cite{masi16faceDuplicated}, 3D facial models \cite{gecer2019ganfit},
signature
\cite{popel07signatureSynth},
or handwriting synthesis
\cite{choi2003generationHandwritten}.

\textbf{Combination methods}: in this case, a pool of bona fide units,
such as $n$-phones in speech (isolated or combination of sounds) or
$n$-grams in handwriting (isolated or combination of characters), is
used as input for the algorithm, which combines or concatenates them
to form the synthetic samples. As in the previous case (i.e.,
transformed samples), the synthetic sample corresponds to the same
subject as the initial units. This is the approach followed by most
speech
\cite{black95speechUnits},
signature \cite{popel07signatureSynth}, and handwriting
\cite{lin07SyntheticHW} synthesisers.


\textbf{Morphing methods}: these algorithms are a special type of
transformation methods, which aim at converting a bona fide sample
belonging to one subject (i.e., source) into a bona fide sample belonging
to a second subject (i.e., target), usually with the intention that
the resulting synthetic sample can be positively matched to both of
the previous identities.

Numerous efforts have been directed towards this research field
within the speaker recognition community
\cite{stylianou2009voiceMorphing}, stemming from the initial proposals
from the late 80's and early 90's
\cite{abe1988voiceConvQuant,valbret1992voiceConvPsola} and leading
to the very recent Voice Conversion Challenge 2016
\cite{voiceConversionChallenge2016}. Analogously, the feasibility of generating synthetic faces which can
be positively matched to both the source and the target subjects has
been demonstrated in
\cite{scherhag17morphingScanned}.

\textbf{Generation methods}: these methods are based on generative
models and follow a two-step approach. First, a model of the
biometric characteristic is created from a database of bona fide samples.
In a second stage, new fully synthetic identities following the
underlying distribution of the training set are generated sampling
the constructed model. Additionally, multiple samples of the
synthetic identities can be generated by any of the procedures for
creating transformed samples (explained above). This approach has
been followed to generate synthetic individuals for biometric
characteristics such as iris
\cite{shah06synthIris},
fingerprint \cite{Cappelli03SfingeHandBook}, face
\cite{karras17GANface}, speech
\cite{deleon2010syntheticSpeech},
mouth \cite{du02mouthSynthesis}, handwriting
\cite{plamondon1998generation},
signature
\cite{galbally12PRssiggeI},
mouse dynamics \cite{nazar08invBioMouse},
or keystroke dynamics \cite{rashid09invBioKeystroke}.

In addition to those works, research in deep learning has made tremendous
progress in recent years in the area of generative models such as Generative
Adversarial Networks (GANs) \cite{Goodfellow-GANs-NIPS-2014},
Variational Autoencoders (VAE) \cite{kingma14variationalAutoencoders}, or autoregressive models
\cite{oord16recurrentNN}. In these works, a generator is trained to produce
synthetic images that resemble bona fide images as much as possible. They are fed to a discriminator,
which will try to classify an image as synthetic or real. At the end of the training stage, the generator
should be able to produce high quality synthetic images, which will be classified as bona fides. Only
this part of the system is retained for later use for image synthesis. In addition to the 
generation of generic facial or iris images \cite{karras17GANface,yadav2019synthIrisRaSGAN}, so-called \lq\lq masterprints'' can be also 
successfully produced with GANs \citep{bontrager2017deepmasterprint}. Such fingerprints,
synthetically generated, can be matched to a high number of different bona fide fingerprints,
belonging to different subjects and thus representing different identities. And last but not least, fingerprints representing a particular identity can also be reconstructed using GANs \cite{kim2018reconstructionFpGAN}. These models have therefore already found a wide
range of applications in different fields, and are expected to
provide a big boost to biometric synthetic generation in the near
future \cite{karras17GANface,kohli17GANiris,oord16GANaudio,gecer2019ganfit,bontrager2017deepmasterprint}.


\textbf{Inversion methods}: also referred to in the literature as
\emph{inverse biometrics}. These methods take as input a bona fide
template and, using some kind of reverse engineering process, they
reconstruct a synthetic biometric sample, which matches the stored
biometric reference according to one or several biometric recognition system(s). 
In other words, they take advantage of the information
conveyed in bona fide biometric templates to gain some knowledge of the
underlying biometric information, hence violating the privacy of the
owner. Such methodologies have already been applied to fingerprint
\cite{hill01AttackDB,Cappelli07PAMIReconstruction,kim2018reconstructionFpGAN}, iris
\cite{venugopalan11irisImageFromBinary,Galbally2013}, handshape
\cite{marta13handInvJournal}, face
\cite{adler03hillfaces,Mai-DeepFaceReconstruction-PAMI-2018}, or
handwriting
\cite{kummel2010handwritingHashInvBio,kummel2010handwritingInvBio}.

As it may be observed in Table~\ref{tab:apps}, the main application
field of this particular kind of synthetic data is vulnerability
analysis: the reconstructed samples are used to impersonate the
subject to whom the bona fide template belongs. Since these algorithms
are the main focus of the present survey, they are highlighted in
blue in Fig.~\ref{fig:SynthBioClassif}, and will be analysed in
detail in the following sections.


In summary, the works on synthetic biometric samples generation
referred in this section have shown how wide this particular field
of research is, and the numerous applications that they cover. In
the following, we will focus on the main purpose of the survey, that
is, the analysis of ``inverse biometrics'', as it is probably one of
the most challenging areas from a security and privacy perspective.

\section{The Threat: Inverse Biometrics Methods}
\label{sec:invBio}

Based on the discussion so far, from a security and privacy
perspective, the distinctive feature of inverse biometric methods
that make them unique with respect to the other four types of
synthetic generation methods described in the previous section, is
that inversion approaches are able to reveal additional information
from the individual, beyond the one that the potential attacker
already possesses.

In the case of the first three types of methods (i.e.,
transformation, combination, and morphing), in order to generate
synthetic samples, the attacker needs to obtain first bona fide biometric
data belonging to the corresponding subjects. Therefore, the new
synthetic data generated does not disclose any further information
from the individuals.

\begin{table*}[t]
\begin{small}
\begin{center}

\caption{Summary of some key inverse biometric approaches.
``Knowledge'' refers to the type of knowledge required to carry out
each method (see Sect.~\ref{sec:invBio} and
Fig.~\ref{fig:SynthBioClassif}). ``Scenario'' refers to the attacking
scenario evaluated on the corresponding article (see
Sect.~\ref{sec:eval}), which yielded the specified ``IAMR''
on the mentioned ``Database''. Whenever an identification instead of
a verification system was evaluated, the Attack is denoted as
``Id''. ``BTP'' stands for the biometric template protection scheme
used (if any).}
\label{tab:invBio} \centering \hspace*{-0.8cm}
\begin{tabular}{llccccc}
\toprule
\textbf{Knowledge} & \textbf{Characteristic} &\textbf{BTP} & \textbf{Ref.} & \textbf{Scenario} & \textbf{IAMR} & \textbf{Database} \\ \midrule

\multirow{10}{*}{\shortstack{Template\\Format}}  &  \multirow{10}{*}{Fingerprint} & & \multirow{2}{*}{\cite{hill01AttackDB}} & \multirow{2}{*}{Id, Sc. 1 $N = 1$} & 100\% & FVC2000 \\
 & & & & & Rank 1 & 110 subj. \\ \cline{4-7}
&  & &\multirow{2}{*}{\cite{Cappelli07PAMIReconstruction}}  & \multirow{2}{*}{Sc. 1 $N > 1$}& $>90\%$ & FVC2002-DB1  \\
  & &  & & & 0.1\% FMR& 110 subj. \\ \cline{4-7}
&  & & \multirow{2}{*}{\cite{galbally09FPsPRLda}}  & \multirow{2}{*}{Sc. 1 $N > 1$}& $>99\%$ & FVC2006  \\
  & & & & &  0.1\% FMR& 140 subj. \\ \cline{4-7}
  &  & & \multirow{2}{*}{\cite{ross07RecFingerprints}}  & \multirow{2}{*}{Id, Sc. 1 $N = 1$}& $>23\%$ & \multirow{2}{*}{NIST-4f}  \\
  & & & & &  Rank 1 & \\ \cline{4-7}
   &  & & \multirow{2}{*}{\cite{kim2018reconstructionFpGAN}}  & \multirow{2}{*}{Sc 1., Sc. 2 $N = 1$}& 98-84\% & CVLab  \\
  & & & & &  FMR = 0.1\% &  380 subj.\\
  \midrule
\multirow{12}{*}{\shortstack{Similarity\\Scores}} & \multirow{6}{*}{Face}  & & \multirow{2}{*}{\cite{adler03hillfaces}} & \multirow{2}{*}{-} & \multirow{2}{*}{-} & \multirow{2}{*}{FRS}  \\
 & & & & & &\\ \cline{4-7}
& & &\multirow{2}{*}{\cite{adler04quantizedFaces}}  & \multirow{2}{*}{Sc. 1 $N = 1$}& $>95\%$ & NIST Mugshot  \\
&  & & & & 1\% FMR& 110 subj. \\ \cline{4-7}
 & & &\multirow{2}{*}{\cite{Mai-DeepFaceReconstruction-PAMI-2018}}  & \multirow{2}{*}{\shortstack{Id\\Sc. 1 $N \ge 1$}}& 39\%-96\% & \multirow{2}{*}{\shortstack{LFW, FRGC\\FERET}}  \\
&  & & & & 0.1\% FMR &  \\ \cline{2-7}
&  \multirow{4}{*}{Iris} & & \multirow{2}{*}{\cite{Galbally2013}} & \multirow{2}{*}{All}& 94\% & BioSecure  \\
 & & & & &0.01\% FMR & 210 subj. \\ \cline{4-7}
& &  & \multirow{2}{*}{\cite{rathgeb10IrisHC}} & \multirow{2}{*}{Sc. 1 $N = 1$}& 100\% & CASIAv3 INt.  \\
& & & & &MS $> 0.9$& 249 subj. \\  \cline{2-7}
  &  \multirow{2}{*}{Handshape} & & \multirow{2}{*}{\cite{marta13handInvJournal}}& \multirow{2}{*}{All}  & 50-90\% & UST DB \\
& & & & &0.1\% FMR & 564 subj. \\
\midrule
\multirow{2}{*}{\shortstack{Distance\\Function}} &  \multirow{2}{*}{Face} & & \multirow{2}{*}{\cite{mohanty07InvBiomFace}}  & \multirow{2}{*}{Sc. 1 $N = 1$}& $>72\%$ & FERET \\
 & & & & & 1\% FMR & 1196 subj.\\ \cline{1-7}

\multirow{16}{*}{\shortstack{Feature\\Extraction}}&\multirow{6}{*}{Face} & & \multirow{2}{*}{\cite{marta12FaceUphill}}  & \multirow{2}{*}{Sc. 2 $N > 1$} & 100\% & BioSecure \\
  & & & & &  0.01\% FMR& 210 subj. \\ \cline{4-7}
  &  & & \multirow{2}{*}{\cite{galbally10PRfaceHC}}  & \multirow{2}{*}{Sc. 2 $N > 1$} & 99-100\% & XM2VTS \\
  & & & &  & 0.1\% FMR& 295 subj. \\ 
   \cline{4-7}
  &  & & \multirow{2}{*}{\cite{zhmoginov16recFacesCNN} }  & \multirow{2}{*}{-} & \multirow{2}{*}{-} & \multirow{2}{*}{-} \\
  & & & &  & &  \\ \cline{2-7}
& \multirow{6}{*}{Iris}   & & \multirow{2}{*}{\cite{venugopalan11irisImageFromBinary}} & \multirow{2}{*}{Sc. 1 $N = 1$}& $>96\%$ & NIST ICE 2005   \\
 & & & & &0.1\% FMR & 132 subj. \\ \cline{3-7}

    & &  \multirow{2}{*}{\shortstack{Fuzzy\\commitment}} & \multirow{2}{*}{\cite{rathgeb2011statisticalAttackFuzzy}} & \multirow{2}{*}{Sc. 1 $N = 1$}& \multirow{2}{*}{-} & \multirow{2}{*}{\shortstack{CASIA v3\\IIT Delhi}}  \\
& & & & & & \\ \cline{3-7}
  & & \multirow{2}{*}{Bloom Filters}   &  \multirow{2}{*}{\cite{bringer15secAnalysisBF}} & \multirow{2}{*}{Sc. 1 $N = 1$}& \multirow{2}{*}{-} & IITD Iris  \\
 &  & & & & & 224 subj. \\  \cline{2-7}
 & \multirow{2}{*}{Handwritting} & \multirow{2}{*}{BioHashing}   &  \multirow{2}{*}{\shortstack{\cite{kummel2010handwritingHashInvBio}\\ \cite{kummel2010handwritingInvBio}}} & \multirow{2}{*}{Sc. 1 $N = 1$}& \multirow{2}{*}{$<70\%$} &  \multirow{2}{*}{5 subj.}   \\
& &   & & & & \\ \cline{2-7}
  & \multirow{2}{*}{Fingerprint} & \multirow{2}{*}{Fuzzy vault}   &  \multirow{2}{*}{\cite{chang2006findingChaff}} & \multirow{2}{*}{-}& \multirow{2}{*}{-} &  \multirow{2}{*}{-}   \\
&  & & & & & \\ \cline{2-7}
  &\multirow{2}{*}{General} & \multirow{2}{*}{\shortstack{Fuzzy\\commitment}}   &  \multirow{2}{*}{\cite{stoianov2009securityBE}} & \multirow{2}{*}{-}& \multirow{2}{*}{-} &  \multirow{2}{*}{-}   \\
 & & & & & \\
\bottomrule
\end{tabular}\vspace{-0.5cm}
\end{center}
\end{small}
\end{table*}
\afterpage{\clearpage}

In the case of generative methods, researchers have shown that
through the so called \emph{membership inference attacks} it is
possible to determine if a given bona fide sample was used to train the
model \cite{shokri17membershipInference}.
While this represents a potential source of information leakage that
can affect privacy, as in the previous three cases, the attacker
needs to be already in possession of the bona fide sample.

On the contrary, the main goal of inverse biometric methods is the following:
starting from a theoretically secure representation of the subject's
biometric characteristics (i.e., the biometric template), produce a
synthetic biometric sample that can be positively matched to the one
that originated it. Therefore, these methods provide the attacker
with sensible biometric information that was not previously known by him
(i.e., the biometric sample). The reversed engineered samples can
then be used to impersonate a particular subject launching
masquerade attacks \cite{Galbally2013} or even presentation attacks
\cite{galbally09FPsPRLda} (see Fig.~\ref{fig:gummyFingers}).


In order to present all the works related to inverse biometrics in a
meaningful manner, this section follows a four-group categorization
according to the type of knowledge required by each method to be
successful. This knowledge is a key parameter in order to evaluate
the risk posed by each approach, and also serves to establish a more
objective comparison among them. A lower knowledge of the system to
be attacked is easier to be obtained, and therefore implies a higher
threat. As shown in Fig.~\ref{fig:SynthBioClassif}, the four groups
considered in our classification are (from a lower to a higher
knowledge level): 1) Knowledge of the template format; 2) knowledge of the similarity scores; 
3) knowledge of the similarity scores and the comparison function; 4) knowledge of the feature extraction method.

In the following sections, we introduce the inverse biometrics
methods proposed in the literature in terms of the aforementioned
types. A summary, including the experimental setup and performance rate
of the algorithms in terms of the Inversion Attack Match Rate (IAMR, see Sect.~\ref{sec:eval} for further details on
the evaluation methodology and metrics), is shown in
Table~\ref{tab:invBio}.

It may be argued that the reconstruction approaches considered in
this article can be successful only when the reference template is
compromised. Even if it may be difficult, it is still possible in
classical biometric systems where the enrolled templates are kept in
a centralised database. In this case, the attacker would have to
access the database and extract the information, or intercept the
communication channel when the stored template is released for
the comparison. However, the threat is increased in Match-on-Card (MoC)
applications where an individual's reference biometric template is
stored in a smartcard that the subject carries with him in order to
access the system. Such applications are rapidly growing due to
several appealing characteristics, including scalability and privacy
\cite{bergman08mocAdvancesBiom}. This makes MoC systems potentially
more vulnerable to the reconstruction algorithms described in this
article, especially when the biometric data is stored without any
type of encryption \cite{NIST-PIV}, or printed in the clear on
plastic cards as 2D barcodes \cite{Seafarers}.
%

On the other hand, even if access to centralised databases is
theoretically more difficult to obtain, an eventual attacker would
be able to compromise not one but numerous biometric templates. Big
data leakages of this sort have already happened over the last five
years\footnote{\url{https://www.wired.com/2015/09/opm-now-admits-5-6m-feds-fingerprints-stolen-hackers/}}
\footnote{\url{https://www.privacyinternational.org/node/342}}.

\subsection{Knowledge Required: Template Format}
\label{sec:invBio:temp}

In many cases, the format
of the templates is known or can be generated with accompanying
SDKs. For instance, the use of standardised templates, which allow
further compatibility across systems and applications, can also
entail a security drawback: an eventual attacker can use this public
knowledge to reconstruct ``valid'' templates and launch attacks on
the system. 

As already mentioned, the first work that addressed the problem
posed by inverse biometrics was carried out by Hill
\cite{hill01AttackDB}. In this work, a general scheme for the
reconstruction of biometric samples is proposed, consisting in four
successive steps, where only knowledge of the templates format
stored in the database is required. 


The most challenging step defined in the article is devising a
method for reconstructing digital samples given only the stored
templates. In that work, a particular case study on
minutiae-based fingerprint templates is presented, based on three
consecutive steps: 1) fingerprint shape estimation, 2)
orientation field creation, and 3) ridge pattern synthesis. 

A similar approach for the generation of fingerprint samples from
standard minutiae-based fingerprint templates was proposed in
\cite{Cappelli07PAMIReconstruction}. Since the templates follow the
corresponding ISO standard \cite{ISO-IEC-19794}, the format is known
to the attacker. This raises a new concern regarding the use of
standards: on the one hand, they are necessary as they guarantee
interoperability; on the other hand, they provide a lot of
information to potential adversaries. Such a concern reinforces the
need to protect biometric templates. 
In contrast to Hill's approach, the algorithm in
\cite{Cappelli07PAMIReconstruction} allows to obtain different
synthetic samples from a single template by using
different frequency values on tje last step. Also, the algorithm in \cite{Cappelli07PAMIReconstruction} includes a 
rendering step, in which noise is added to the
``perfect'' reconstructed image, thus yielding more realistic
images. As originally suggested by \cite{hill01AttackDB}, starting
from those synthetic images, gummy fingers can be generated to
carry out presentation attacks as demonstrated in
\cite{galbally09FPsPRLda} (see Fig.~\ref{fig:gummyFingers}).

A different approach is followed in \cite{ross07RecFingerprints} to
reconstruct fingerprint images, in which, contrary to \cite{hill01AttackDB,Cappelli07PAMIReconstruction}, no iterative technique is considered. The only knowledge required are the minutiae positions and orientations. Following the example of \cite{Cappelli07PAMIReconstruction}, this work also added a rendering step in order to generate more realistic fingerprints.

Finally, Kim \textit{et al.} \cite{kim2018reconstructionFpGAN} have used GANs for the generation of fingerprint images based only on the minutiae positions and orientations, as in \cite{ross07RecFingerprints}. As described in Sect.~\ref{sec:synthSamp}, the generator and discriminator networks are trained together, but at a latter stage only the generator is necessary to produce synthetic fingerprints. Contrary to the some of the aforementioned works, no rendering step is necessary in this case to produce realistic images: the GAN is able to synthesise them.


\subsection{Knowledge Required: Similarity Scores}
\label{sec:invBio:scores}


In this case, the attacker only
needs to use the biometric system as a black box: he feeds probe
images to the system and receives the similarity score with respect
to the reference as a feedback. It is true that such information is
not always available in commercial systems. Nevertheless, for some
biometric characteristics, like the iris, for which the systems are
mostly based on a single approach (i.e., Daugman's algorithm), this
knowledge is readily available from opensource systems as well. In fact, the template reconstructed with
such public system can be even used to fool other commercial systems
for which the information is not available.

The method proposed in \cite{adler03hillfaces} for the
reconstruction of face samples from Eigenface based templates relies
on a hill-climbing optimization of synthetic face images. The
authors use the similarity score between the synthetic images and
the stored template as feedback to improve the synthetic
reconstruction. A more efficient hill-climbing technique is proposed
in \cite{adler04quantizedFaces}, where each quadrant of the
synthetic face image is independently optimised even if only
quantised scores are shared by the verification system (as
recommended by the BioAPI specification \cite{bioapiEspec}).

Rathgeb and Uhl proposed in \cite{rathgeb10IrisHC} a different
inverse biometrics method for iris templates based on a
hill-climbing algorithm using synthetic samples. In spite of the
high dependency of the positive verification of iris textures on the
feature extraction algorithm, the authors describe a general
approach for synthetic iris textures generation where no knowledge
about the extracted features is required. This method builds upon
the fact that most iris recognition algorithms share a common
characteristic: they tend to average pixels in a block-wise manner.

A different scheme was proposed by Galbally \textit{et al.}
\cite{Galbally2013} to reconstruct iris patterns from their
corresponding iris binary templates, using a probabilistic approach
based on genetic algorithms, which is able to reconstruct several
different images from the same iris template. The approach needs to
have access to a comparison score which does not necessarily have to
be that of the system being attacked. This way, the reconstruction
approach is somewhat independent of the comparator or feature extractor
being used. 
The authors showed that the
algorithm can successfully bypass black-box commercial systems with
unknown feature-extraction algorithms.

Similarly, Gomez-Barrero \textit{et al.} proposed in
\cite{marta13handInvJournal} a probabilistic inverse biometrics
method based on a combination of a handshape images generator and an
adaptation of the Nelder-Mead simplex algorithm,
which had been previously used to recover face images in
\cite{marta12FaceUphill}. 

More recently, Mai \textit{et al.}
\cite{Mai-DeepFaceReconstruction-PAMI-2018} analysed the
vulnerabilities of state of the art face recognition systems based
on Convolutional Neural Networks (CNNs). Even if it is argued that
only knowledge about similarity scores between synthetic probe and
bona fide reference templates is required, it should be noted that the
templates are required to be the output of CNNs (therefore, some
information about the template format is needed). 

\subsection{Knowledge Required: Similarity Score and Comparison Function}
\label{sec:invBio:dist}

In this case, the impostor
also needs to know the topology of the comparison function in order
to extract additional information to the plain similarity scores.
Therefore, this approach is more challenging for the attacker than
the previous one, and has attracted less attention in the literature.

Compared to \cite{adler03hillfaces}, where only the similarity scores were needed, Mohantly \textit{et al.}
reconstruct face images in \cite{mohanty07InvBiomFace} assuming
access to the similarity scores between a pool of bona fide face images
and the face to be reconstructed. Furthermore, knowledge of the
comparison function used by the particular face verification system
is required. In this method, the authors model the face sub-space
with an affine transformation. In order to reconstruct a particular face enrolled in
the system, the distances from the pool of bona fide images to the
attacked face are used to compute the point in the affine subspace
that corresponds to the attacked identity.


\subsection{Knowledge Required: Feature Extraction}
\label{sec:invBio:feat}


Some algorithms
require knowledge of this module in order to reverse-engineer it and
reconstruct biometric samples from an optimised template. This is
thus the most challenging scenario in terms of knowledge, since the
attacker needs to be in possession of many details about the system, which, for
commercial systems, is in most cases very difficult to obtain.
However, coming back to the iris case, such attacks can still pose a
severe threat.

To this class belongs for instance the method by Venugopalan and
Savvides, which reconstructs iris samples from the iris binary templates
in \cite{venugopalan11irisImageFromBinary}, where a reversed version of the Gabor function used to extract the binary templates is used together with a pool of bona fide iris samples to generate the reconstructed iris pattern.

In a similar manner to the approach proposed in
\cite{adler04quantizedFaces}, where only knowledge of the similarity
scores was required, face images are recovered from Eigenface and
Gaussian Mixture Models (GMM) parts-based systems in \cite{galbally10PRfaceHC}. In this case,
a Bayesian hill-climbing algorithm is used to optimise the feature
vectors instead of the input samples. The optimised templates are
reverse-engineered to obtain the final synthetic images.
Analogously, Gomez-Barrero \textit{et al.} reconstruct face samples
from Eigenface systems in \cite{marta12FaceUphill} by means of the
downhill simplex algorithm. It
should be noted that the hill-climbing attacks described in those
works \cite{galbally10PRfaceHC,marta12FaceUphill} can be launched on
any system as long as the adversary has access to: 1) scores and
2) template format. However, the reconstruction process to
recover the face image only works if the feature extractor uses
Eigenfaces (i.e., knowledge of the feature extractor required). Face
images were also recovered from their eigenface representations in
\cite{mignon13recFacesRBFregression} based on Radial Basis Functions
(RBF) regression. In this last case, the exact mapping function
extracting the templates needed to be known.

Recently, in \cite{zhmoginov16recFacesCNN}, the authors showed that
the hill-climbing attacks based on gradient ascent which require the
access to multiple successive similarity scores can be instead
replaced by a single-attempt reconstruction method based on CNNs. In
this case, however, knowledge of the feature extraction process is
required. 
\section{Evaluating the Threat: Assessment of Inverse Biometric Methods}
\label{sec:eval}

One of the major open issues that still need to be addressed in the field of inverse biometrics is the development of a common methodology to evaluate the inversion algorithms and the risk that they pose to the attacked applications. Although an initial attempt of an evaluation framework was proposed and followed in \cite{Galbally2013,marta13handInvJournal}, most of the works published in this field use different protocols and metrics for the evaluation. This makes the task of comparing the results obtained in each article very difficult, if possible at all.

In order to bridge this existing gap, in the present section, we propose a new evaluation methodology for inverse biometric algorithms which partly builds on the results of \cite{Galbally2013,marta13handInvJournal}, but that takes into consideration the general vulnerability framework for biometrics that is being developed in the ISO/IEC 30107 standard \cite{ISO-IEC-30107}. This framework already considers under its umbrella the evaluation of presentation attacks \cite{hadid2015padeval} and morphing attacks \cite{ulrich17morphingMetrics}, and we believe that it provides the necessary tools to also include in it the evaluation of inversion attacks.

\subsection{The Evaluation Methodology}
\label{sec:eval:description}

\begin{figure*}[t]
\centering
\begin{minipage}[b]{0.99\linewidth}
  \centering
  \centerline{\includegraphics[width=.99\linewidth]{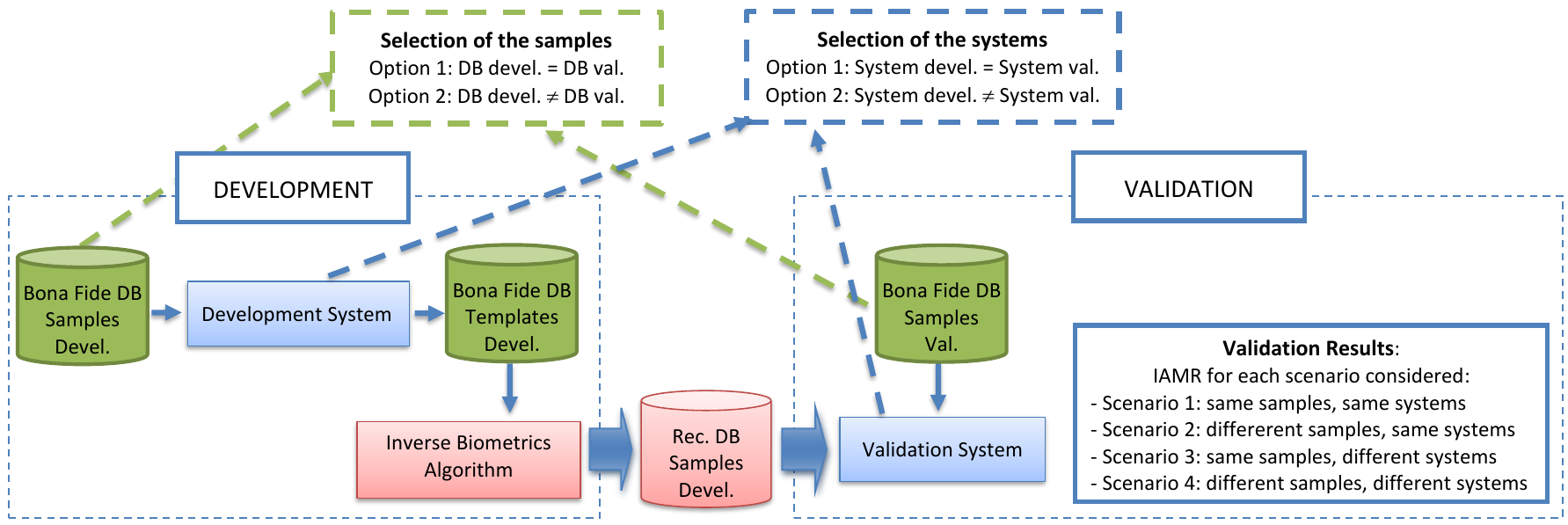}}
\end{minipage}
\caption{Two-stage experimental protocol proposed for the evaluation
of the threat posed by inverse biometric algorithms: 1) in the
development stage, the reconstructed database is generated from the
templates produced using a development system, and 2) in the
validation stage, the privacy threat posed by the reconstructed
samples is evaluated launching attacks on one or more validation
systems. In
the figure, bona fide databases are depicted in green, and synthetic
databases in red.} \label{fig:expProtocolGeneral} \vspace*{-0.5cm}
\end{figure*}

As presented in Sect.~\ref{sec:invBio}, inverse biometric algorithms
have been traditionally developed within the biometric security
field as attacking methods. Accordingly, their assessment has been
addressed in most cases following the principles of vulnerability
evaluations. In this particular case, the objective being to
determine the threat posed by the synthetic reconstructed samples on
biometric systems.

The attacking scenario usually considered can be summarised as
follows. For a given biometric system, an attacker retrieves the
template of a particular subject, reconstructs the corresponding bona fide sample by means of
some of the inverse biometric methods described so far, and tries to
illegitimately access the system using it.

Taking this context into account, the main goal of the evaluation is
to determine the success rate of an eventual \emph{inverse biometrics} or \emph{inversion}
attack such as the one described above. That is, what is the
probability that a synthetic reconstructed sample is positively
matched to a bona fide template of the legitimate user? This assessment
will also: 1) determine the performance of the reconstruction
approach (i.e., how good are the synthetic samples produced by a
given method?); 2) allow benchmarking the efficiency of different
reconstruction approaches (i.e., what inverse biometric method is
more efficient?); and 3) give an estimation of the reversibility
level of the templates (i.e., to what extent is it feasible to
reverse engineer them?).


It should be noted that the previous process can be interpreted from
two points of view, depending on whether the evaluator is more
interested by 1) the assessment of the performance of the inverse
biometric algorithm or by 2) the assessment of the vulnerability of
a given system to this threat. These two dimensions of the same
problem are not independent. On the contrary, they are fully
interrelated and, in many cases, the same metrics can be used for
the assessment of both perspectives. The methodology presented in
this section is general and can be used to evaluate both. In subsection~\ref{sec:eval:details}, some further discussion is given on how to adapt
the methodology in order to put the stress on the inversion algorithm
or on the recognition system.

The proposed evaluation protocol, as depicted in Fig.~\ref{fig:expProtocolGeneral}, is divided into a development and a validation stage.

 \textbf{Development}. This first stage has a two-fold objective:
1) if necessary, train any module of the reconstruction algorithm;
and 2) generate the synthetically reconstructed dataset that
will be used in the validation stage. It is in this stage when the
inverse biometric method to be evaluated will be used to generate
the synthetic samples. Applying the method to each template of the
targeted database, a new synthetic database, depicted in red in
Fig.~\ref{fig:expProtocolGeneral}, is generated, comprising at least
one synthetic sample generated from each bona fide template in the
bona fide development database (depicted in green).

\textbf{Validation}. Once the synthetic samples have been
generated, the objective is to estimate the probability that the
reconstructed samples can be used to successfully bypass a
recognition system. That is, the validation system is the one whose security / reversibility is being evaluated.
To this end, the synthetically reconstructed
samples are presented to the validation biometric system to determine if they are
positively matched to the corresponding bona fide samples or reference templates. By evaluating the number 
of times the attack is successful, it is possible to determine: 1) the reversibility of the templates produced by the validation system; 
2) the vulnerability of the validation system to this type of attack; and 3) the efficiency of the inversion method.

It should be noted that, from a general perspective, different
databases and systems might be used at the development and validation stages, as
depicted in Fig.~\ref{fig:expProtocolGeneral} (see
Sect.~\ref{sec:eval:details} for further discussion regarding the
use of the same or different datasets / systems at both stages). Please also recall
that the level of knowledge required by the inverse biometrics
method defined in Sect.~\ref{sec:invBio} is referred in all cases
to the development system, which is the one producing the templates
to be reversed engineered.


\subsubsection{Selection of the samples}
First, it is important to be aware that the \emph{subjects} present in
the ``Bona Fide DB Samples Development'' and in the ``Bona Fide DB Samples
Validation'' are always the same. However, depending on the scenario being considered, the
\emph{samples} belonging to these subjects contained in each of the
DBs may be different. Therefore, as it is depicted in Fig.~\ref{fig:expProtocolGeneral}, two different options may be considered:

 \textbf{Option 1}. The bona fide sample being attacked is the
\emph{same} bona fide sample that produced the template from which
the synthetic samples were reconstructed. This is the most \emph{basic} type of attack and should be included
in any evaluation. It provides a measurement of the efficacy of
the inverse biometrics algorithm to accurately reconstruct the
bona fide sample from which a template was extracted, which at the
same time can be used as an estimation of the reversibility of the
templates. Therefore, this option also gives a first indication of the
vulnerability of the system to the attack.

\textbf{Option 2}. The bona fide sample being attacked is a
\emph{different} sample (of the same subject) to the one that
produced the template from which the synthetic samples were
reconstructed. This is probably the most \emph{realistic} security scenario,
where the attacker is able to obtain a template which does not come
from the sample that will be used by the system for recognition
purposes. Therefore, this attack complements the information
produced by option 1 regarding the vulnerability of the system.

In both cases, $N \ge 1$ reconstructed samples are generated from one bona fide
template, and subsequently compared to one bona fide sample or reference template.
This represents the most likely attack strategy analysed in
other related vulnerability studies
\cite{Cappelli07PAMIReconstruction}, where the template of a
legitimate subject in the database is compromised and the intruder
reconstructs multiple samples to try and break the system. The
attacker will gain access if any of the reconstructed samples
results in a positive match. It also implies that the attacker can
potentially gain access to the system multiple times, without being
detected due to the use of the exact same sample.

It should be however noted that
$N > 1$ implies that the reconstruction method
is able to produce different samples from one given template, which
is usually the case for non-deterministic algorithms. This case can help to
further analyse: 1) the ability of the inverse biometrics algorithm
to generalise and produce samples within the intra-variability of a
given individual and not to overfit to one specific sample; 2) the
vulnerability of the system to be bypassed by an eventual attack
carried out by several different synthetic samples (higher threat).


\subsubsection{Selection of the systems}

As depicted in Fig.~\ref{fig:expProtocolGeneral}, 
the same or different recognition systems may be used at each stage of the evaluation methodolgy. Just as a
reminder: the system at the development stage produces the templates
from which the reconstructed samples are generated, whereas the system at
the validation stage compares the synthetic reconstructed samples to
the bona fide samples. With this in mind, whether to use the same or
different systems mainly depends on what is the target of the
evaluation: the inverse biometrics algorithm or the recognition
system.

 \textbf{Option 1}: Assessment of the Inverse Biometrics Algorithm. If the main goal
is to assess the efficiency of the inverse biometrics algorithm to
reconstruct samples, using \emph{different} systems at both stages will
lead to a more complete evaluation of the threat posed by the
inverse biometrics method, since we are reducing the dependency on
the biometric comparator which could justify the success of the
reconstruction. That is, an evaluation with several systems will
show the level of generality of the method and if it is designed to
1) target one specific system (this would be the case of considering
the same system at development and validation), or if, on the
contrary, 2) it poses a risk to a whole range of systems (i.e., the
reconstructed samples are not only recognised by the system that
produced the templates but by different ones).

\textbf{Option 2}: Assessment of the recognition system. If the main goal is to
assess the security of a given recognition system (i.e.,
reversibility of its templates), then it would only be meaningful to
consider the \emph{same} system (the one being evaluated) both at
development and validation.

In summary, option 1 is more focused on evaluating the
inverse biometrics algorithm and its ability to reverse engineer
templates (i.e., also a metric for template reversibility), while
option 2 is more focused on evaluating the vulnerability
of the validation recognition system to an attack carried out with
synthetic samples (i.e., how likely is it that the system will be
bypassed).

\subsubsection{Inversion metric}
Depending on the selection of the samples and the systems, there are four possible evaluation scenarios:
\begin{itemize}
\item \textbf{Scenario 1}: Same samples, same systems
\item \textbf{Scenario 2}: Different samples, same systems
\item \textbf{Scenario 3}: Same samples, different systems
\item \textbf{Scenario 4}: Different samples, different systems
\end{itemize}

The performance of the inversion attack under each
scenario can be measured in terms of its Inversion Attack Match Rate (IAMR) for
a given operating point of the biometric system. The IAMR is accordingly defined as the expected
probability that a reconstructed sample gains access to the system at a given operating point as follows:

\begin{equation}
\mathrm{IAMR} = \frac{1}{M}\sum_{m = 1}^M \left\lbrace \max_{n \le N}\left\lbrace S_m^n\right\rbrace > \delta \right\rbrace\label{eq:IAMR}
\end{equation}
where $M$ is the number of subjects being attacked, $N$ the number of reconstructed samples per bona fide sample, $S_m^n$ the similar score of the $n$-th reconstructed sample of the $m$-th subject, and $\delta$ is the verification threshold. The $\mathrm{max}$ function represents the aforementioned fact that an attacker may reconstruct multiple synthetic samples and will success in his goal if one of the samples is positively matched to the bona fide template.

This measure gives an estimation of how dangerous a particular
attack is for a given biometric system: the higher the IAMR, the
bigger the privacy threat. Or, in other words, for the case of
inverse biometrics, the more reversible the templates. As in other
biometric vulnerability evaluations, the success of an attack is
highly dependent on the False Match Rate (FMR) of the system: the
higher the FMR, the easier it is for an impostor (i.e., synthetic
sample) to be accepted. As a consequence, it should always be
specified for a given evaluation, the operating point at which the
IAMR has been computed. Some operating points typically
used are FMR = 0.1\%, FMR = 0.05\%, and FMR = 0.01\%, which,
according to \cite{ANSI-MinFMR}, correspond to a low, medium and
high security application, respectively. For completeness, systems
should be also tested at very high security operating points (e.g.,
FMR $\ll$ 0.01\%).


\subsection{Notes on the Evaluation Methodology}
\label{sec:eval:details}

The methodology described above is general and needs to be
adapted to the context of each particular evaluation on a case by
case basis. In particular, please be aware that in order for the IAMR to be informative, the protocol followed in the evaluation should be clearly explained. That means that each evaluation should define the pair of development / validation systems (that may or may not be the same) and the samples scenarios considered (same / different development and validation samples). Therefore, it is very important to define which of the four scenarios covered by the methodology is being considered in the evaluation:
\begin{itemize}
\item Scenario 1: Same samples / Same systems in development and validation. This scenario simulates the case in which the attacker is in possession of the biometric reference template used by the system he wants to break. Therefore, it constitutes the lowest security risk.
\item Scenario 2: Different samples / Same systems in development and validation. This scenario illustrates the real case where the attacker is in possession of any biometric template of the subject, not necessarily the one stored as reference in the database. This means that, if two different applications use the same biometric recognition algorithm (i.e., system), and they have different reference templates of a given subject, the attacker CAN break both of them by having access to any template extracted from the subject. Hence, this poses higher security risk than scenario 1.
\item Scenario 3: Same samples / Different systems in development and validation. In this case, the attacker is again in possession of the biometric reference template used by the system he wants to break, but does not have access to the system itself. Therefore, he uses another system to reconstruct the synthetic biometric sample. In contrast to scenarios 1 and 2, if a particular application makes use of an expensive recognition system, the attacker does not need to acquire it. In addition, the inversion algorithm is able to generalise, most likely, to diverse validation systems, thereby increasing the security risked posed by it.
\item Scenario 4: Different samples / Different system in development and validation. This is the most challenging scenario from a security perspective, since the attacker is neither in possession of the particular system he wants to attack, nor the exact reference template stored in the database. That means, that he is eventually able to impersonate a subject enrolled at different applications, maybe using different recognition systems.
\end{itemize}

It is important to notice that these scenarios pose an increasing security risk from 1 to 4, as discussed above. Or, in other words, the inverse biometrics algorithm will pose a smaller privacy threat if it is only capable of successfully reconstructing templates of a unique system (used both for development and validation) and when these templates are also the ones used in validation (scenario 1). The highest security risk is posed when a high IAMR is obtained for different samples and systems (scenario 4). But it should not be forgotten that the amount of information about the system required by the inversion method to be successful also plays a key role in determining the risk level of the algorithm.

As a way to better illustrate the potential of the evaluation methodology, in the next Sect.~\ref{sec:eval:example} we present a practical example of how it can be applied to a real case study and how to report its results.

\subsection{Case Study on Iris Templates}
\label{sec:eval:example}

In this section, we illustrate the use of the evaluation methodology for the reconstruction of iris images starting from their corresponding templates, known as iriscodes. Furthermore, we compare two methodologies, presented in \cite{Galbally2013,venugopalan11irisImageFromBinary}. Please recall that the knowledge required refers to the development system (see Sect.~\ref{sec:invBio}).

The method proposed by Venugopalan and Savvides in \cite{venugopalan11irisImageFromBinary} is evaluated within the following framework:

\begin{itemize}
\item Knowledge required about the development system: template format and feature extraction method.
\item Samples selection: same development and validation samples, with $N = 1$.
\item System selection: same development and validation system (log-Gabor based).
\end{itemize}
Therefore, out of the four scenarios described above, only scenario 1 is analysed in this work.

\vspace{0.2cm}

The method proposed by Galbally \textit{et al.} in \cite{Galbally2013} is evaluated within the following framework:
\begin{itemize}
\item Knowledge required about the development system: template format.
\item Samples selection: both same development and validation samples, and different development and validation samples, with $N = 5$.
\item System selection: fixed development system (open source log-Gabor based), both same and different (commercial off-the-shelf system (COTS) black-box system) validation systems.
\end{itemize}
Therefore all scenarios 1 to 4 are considered in this work.

\vspace{0.2cm}

\begin{table}[t]
\begin{small}
\begin{center}

\caption{Case study evaluation results for the methods described in \cite{Galbally2013,venugopalan11irisImageFromBinary}. The IAMR (see Eq.~\ref{eq:IAMR}) is shown at different operating points and for the scenarios evaluated in each work.}\label{tab:ex} \centering
\begin{tabular}{llccc}
\toprule
& & \phantom{c} &  \multicolumn{2}{c}{Method} \\
& & & \cite{venugopalan11irisImageFromBinary} & \cite{Galbally2013}\\
\midrule
\multirow{3}{*}{Scenario 1} & FMR = 0.1\% & &92.6\% & 100\% \\
& FMR = 0.05\% & &$\approx$89\% & 100\%\\
& FMR = 0.01\% & & $\approx$85\%& 100\%\\ \midrule
\multirow{3}{*}{Scenario 2} & FMR - 0.1\% & & - & 98.7\% \\
& FMR = 0.05\% & & - & 97.9\%\\
& FMR = 0.01\% & & - & 96.5\%\\ \midrule
\multirow{3}{*}{Scenario 3} & FMR = 0.1\% & & - & 96.2\% \\
& FMR = 0.05\% & & - &96.2\% \\
& FMR = 0.01\% & & - & 95.2\%\\ \midrule
\multirow{3}{*}{Scenario 4} & FMR = 0.1\% & & - & 92.8\% \\
& FMR = 0.05\% & & - & 91.4\%\\
& FMR = 0.01\% & & - & 90.9\%\\
\bottomrule
\end{tabular}\vspace{-0.5cm}
\end{center}
\end{small}
\end{table}

The resulting IAMR values for each of the scenarios considered in the two works are summarised in Table~\ref{tab:ex}\footnote{Whereas all results were available in \cite{Galbally2013}, for \cite{venugopalan11irisImageFromBinary}, the values have been extracted from Fig. 7b and Table III in the article, since IAMR values in Fig. 9 for FMR lower than 1\% cannot be discerned. It should be noted, that Venugopalan and Savvides indicate that other experiments have been carried out using different development and validation systems, but the results are not reported. Therefore, a benchmark in this section cannot be carried out for that system selection.}. Since both algorithms have been evaluated under scenario 1, thanks t our approach we can establish a fair benchmark between them. On the one hand, the IAMR values reached by \cite{Galbally2013} are slightly higher (i.e., from 8\% up to 15\%, depending on the operating point of the system). On the other hand, also the knowledge required by Galbally \textit{et al.}'s method is easier to obtain. Therefore, we may conclude that the risks posed by \cite{Galbally2013} to iris recognition systems is higher.

In addition to scenario 1, the same inverison method proposed in \cite{Galbally2013} has been evaluated for the remaining three scenarios. Regarding scenario 2, on which different samples are used for development and validation, we can see a slight decrease in the IAMR. This is an expected effect due to the intra-class variability presented by biometric samples. However, the decrease remains under 4\% even for high security operating points. Therefore, this scenario shows the robustness of the proposed approach to realistic conditions on which the template used for the reconstruction is not identical to the reference template stored in the database.

Finally, in scenarios 3 and 4 we can see a further decrease in the IAMR, due in this case to the use of different development and validation systems. This means that the reconstruction method is optimised for the development system, on which the IAMR achieves the maximum values of 100\% (scenario 1) and 96.5\% to 98.7\% (scenario 2). When the comparator and the feature extractor of the recognition system are changed, these values decrease to 95.2\% to 96.2\% (scenario 3) and 90.9\% to 92.8\% (scenario 4). That is, the chances of breaking the system are reduced by around 5\%, which we may regard as a small decrease. That is, this inversion method is very consistent even for the most challenging scenarios.



\section{Mitigating the Threat: Countermeasures to Inversion Attacks}
\label{sec:btp}

%

The aforementioned works have shown that, contrary to the
traditional belief that the extracted templates did not reveal
enough information to reconstruct the underlying biometric data, it
is indeed possible to recover synthetic biometric samples which are
identified as bona fide subjects by the systems. To tackle this severe
issue, several approaches have been followed, as we will review in
the present section.

In the first place, the BioAPI Consortium \cite{bioapiEspec}
recommends that biometric systems output only \textbf{quantised
similarity scores}. Quantization steps should be as big as possible
but without compromising the systems recognition accuracy. This
countermeasure is an effective way to prevent many of the
reconstruction methods based on hill-climbing algorithms
\cite{adler03hillfaces,Galbally2013,marta13handInvJournal}. This is
due to the fact that those reconstruction methods require feedback
on whether or not the score increases in each step of the algorithm.
However, it has been shown that some hill-climbing attacks are still
robust to this countermeasure \cite{adler04quantizedFaces}. As an
alternative, non-uniform quantization is also evaluated by Maiorana
\textit{et al.} \cite{maiorana15multiHC} as a possible
countermeasure. In their work, a fixed number of quantization levels
is chosen based on the Lloyd-Max quantiser \cite{lloyd1982lsq},
determining the intervals so that the mean-square error (MSE)
between the original and quantised distributions is minimised. The
authors highlight that one of the main advantages of this method is
not only its higher efficiency when compared to uniform
quantization, but also its capabilities to adapt to different
attacking scenarios. While hill-climbing attacks are especially
relevant at low FMR operating points, false acceptance attacks
launched with synthetic images are preferable when the systems works
at a higher FMR. Therefore, a finer quantization should be chosen
for the appropriate range of similarity scores values.

In spite of the security enhancement provided by those score
quantisation schemes, it should be noted that they also lead to a
recognition accuracy degradation. In addition, not all attacks can
be prevented \cite{adler04quantizedFaces,marta12FaceUphill}. Due to
these facts, and with the new and more restrictive privacy
regulations such as the EU General Data Protection Regulation (GDPR)
\cite{EU-Regulation-DataPrivacy-160427}, \textbf{biometric template
protection} (BTP) approaches have been devised to prevent inversion attacks in general, and not only specific methodologies
\cite{Rathgeb11e,patel15CancelableBioSurvey}. In these systems,
unprotected templates are neither stored in the database nor
compared for verification purposes. They are substituted by
protected templates so that, in case of leakage, those references
disclose no biometric information about the subjects, hence
protecting their privacy. To that end, protected
templates should comply with the two major requirements of
\emph{irreversibility} and \emph{unlinkability}:

\begin{itemize}

\item \textit{Irreversibility}: in order to minimise the amount of
biometric information which can be potentially leaked by the
template, it is required that a compromised protected template and
any corresponding auxiliary data cannot be exploited to reconstruct
a biometric signal which positively matches the bona fide biometric
sample (i.e., cannot be exploited by inverse biometric algorithms).

\item \textit{Unlinkability}: in addition to not being reversible,
biometric characteristics should not be matched across systems and
they should be replaceable. That is, given a single biometric
sample, it must be feasible to generate different versions of
protected templates, so that those templates cannot be linked to a
single subject \cite{barrero18generalFrameworkUnlinkability}. This
property guarantees the privacy of a subject when he/she is
registered in different applications with the same biometric
instance (prevents cross-matching or linkage), and also allows
issuing new credentials in case a protected template is stolen.

\end{itemize}

Only fulfilling the previous two requirements is the privacy of the
subject fully preserved. 

To that end, different approaches have been followed, which lead to three main BTP systems categories \cite{tuyls07securityNoisyData,campisi13secPrivacyBio}, namely:1 1) \textit{cancelable biometrics} \cite{patel15CancelableBioSurvey}, 2) \textit{cryptobiometrics} \cite{Rathgeb11e}, and 3) \textit{biometrics in the encrypted domain} \cite{bringer13SMCbiometrics}. Their main advantages and drawbacks are summarised in the following paragraphs.

\textbf{Cancelable biometrics} refer to schemes in which biometric data is obscured with an irreversible transformation of the bona fide samples or unprotected templates \cite{ratha01securityPrivacy,campisi10onlineSignBTP,maiorana08HMMbtp,aragones12signatureUBMBTP}, and verification is carried out in the transformed domain. Such transformations can include the addition of some \lq\lq salt'' to the unprotected template to distort it \cite{BTeoh08b,teoh2006biophasor}. In these systems, the privacy of the subject is preserved at all times (i.e., no biometric data or unprotected templates are used or stored). In addition, such transformations are in most cases time efficient. However, most of these approaches, with only a few exceptions \cite{marta16unlinkBF,marta16generalBF}, lead to a degradation in verification performance. Furthermore, should a protected template be stolen, there is no way to recover the bona fide biometric sample or the unprotected template in order to re-encode it with a new key. As a consequence, in order to re-generate the biometric database, biometric samples need to be re-acquired, with the additional nuisance this fact could pose to the subjects. 

On the other hand, in \textbf{cryptobiometric systems} a key is either bound (i.e., key binding schemes) or extracted (i.e., key generation schemes) from biometric data. In this context, most systems rely on the fuzzy vault \cite{BJuels02a} and the fuzzy commitment \cite{BJuels99a} schemes, which are classified as key binding approaches. Those methods share a common drawback: statistical attacks on this AD, used for verification, can compromise both the security of the system and the privacy of the subject \cite{ignatenko09bioPrivacySec,ignatenko10fuzzyComLeakage}. In addition, cryptobiometric systems usually present a performance degradation with respect to the systems relying on unprotected data, and data needs to be re-acquired in case the templates are compromised. To solve these challenges, optimised versions of the initial fuzzy cryptosystems have been proposed in the literature \cite{bringer2007fuzzyIris,7314608}, and hybrid systems profiting from the advantages of both biometric cryptosystems and cancelable biometrics have been presented \cite{BRINGER200843}.

As an alternative to the aforementioned approaches, secure multiparty computation and homomorphic cryptosystems can be used in order to carry out biometric recognition in the \textbf{encrypted domain}, while obtaining results fully comparable to those yielded by plain data \cite{fontaine13surveyHE,lagendijk13ESPprivacy}. In particular, current approaches to biometrics in the encrypted domain \cite{bringer13SMCbiometrics} are based on Garbled Circuits (GC) \cite{yao86garbledCircuits} and Homomorphic Encryption (HE) \cite{fontaine07surveyHE,fontaine13surveyHE}. Since efficient implementations of HE schemes are very recent, only a few systems have been presented \cite{barni10fingercodeHE,bianchi2010FingerCodeCrypto,osadchy10SCiFI,blanton11irisFpHE,bringer2014gshade,marta2016EncDTW}. Being based on traditional cryptographic protocols, their security and privacy protection capabilities have been more thoroughly tested and are supported by rigurous mathematical proofs. However, their recognition performance does not lie within the state of the art in most cases: even though biometric algorithms that achieve better detection rates are known in the literature, these schemes are much more complex than the representations used in the aforementioned articles \cite{nandakumar15BTPPerformance}, due to the limitation in the number of possible operations that can be performed in the encrypted domain and the additional computational load introduced by them.

Finally, it should be highlighted that for all the aforementioned categories, also multi-biometric systems have been proposed \cite{kelkboom2009MultiAlgBTP,ross13mixingFingerprints,nagar2012MultiCrypto,marta16generalBF,marta16multiHE}. The main goal of these later systems is to further increase both the recognition performance by extracting information of complimentary sources (e.g., different biometric characteristics or feature extraction methods), and at the same time increase the privacy protection, especially in the case a feature level fusion is used.

\section{Discussion: The importance of achieving irreversibility}
\label{sec:privacyCounter}

Over the last years, some large-scale initiatives, such as the
Indian Unique ID\footnote{\url{https://uidai.gov.in/}} or the
SmartBorders
package\footnote{\url{http://ec.europa.eu/dgs/home-affairs/what-we-do/policies/borders-and-visas/smart-borders/index_en.htm}},
have adopted biometrics as their recognition technology. Biometric
systems are also being introduced into the banking sector
\cite{bioBanking}, reaching our smartphones through specific apps
for particular
banks\footnote{\url{https://ingworld.ing.com/en/2014-4Q/7-ing-app}},
through general payments apps such as ApplePay or SamsungPay, or
even with Mastercard's \lq\lq selfie''
payments\footnote{\url{http://www.cnet.com/news/mastercard-app-will-let-you-pay-for-things-with-a-selfie/}}.
Furthermore, biometric
ATMs\footnote{\url{http://www.biometricupdate.com/201301/citibank-launches-smart-atms-with-biometric-capabilities-in-asia}}
are currently being deployed. However, in spite of the wide
acceptance and deployment of biometric recognition systems, some
concerns have been raised about the possible misuse of biometric
data \cite{bustard15privacyEU}. Such concerns can be summarised in
the following questions.

\textit{Do stored templates reveal any information about the
bona fide biometric samples? In other words, are we able to
reconstruct synthetic samples similar enough to those of the
original subject?} The works described in Sect.~\ref{sec:invBio}
have shown that, for a wide variety of biometric characteristics and
systems, it is possible to carry out such a reverse engineering
process. As a consequence, an eventual attacker which manages to
obtain just a template belonging to a certain subject (e.g. the iris
binary template or minutiae template) could be able to reconstruct
the bona fide biometric sample. The attacker can afterwards use it to
illegally access the system, to steal someone's identity, or to
derive additional information from the obtained biometric data,
thereby violating the right to privacy preservation of the subject.
As a consequence, we must ensure the \textit{irreversibility} of the
templates. Sect.~\ref{sec:eval} has presented a protocol to measure
their reversibility.

\textit{Even if templates were irreversible, are my enrolled
templates in different recognition systems somehow related to each
other? Can someone cross-match those templates and track my
activities?} We should not only think about protecting the stored
references in order to make infeasible the inversion
process. With the widespread use of biometrics in many everyday
tasks, a particular subject will probably enrol in different
applications, such as health care or on-line banking, with the same
biometric instance (e.g., my right index finger). The right of
privacy preservation also entails the right not to be tracked among
those applications. If the answer to the previous question is yes,
we are facing an additional privacy issue: an eventual attacker who
gets access to several templates enrolled in different systems could
combine that information and further exploit it to gain knowledge of
how many bank accounts we have or infer patterns in our regular
activity. Therefore, \textit{cross-matching} between templates used
in different applications should be prevented.

\textit{Finally, what if someone steals a template extracted
from my right index finger? Will I be able to use that finger again
to enrol into the system? Has it been permanently compromised?}
Since biometric characteristics cannot be replaced, we should be
able to generate multiple templates from a single biometric instance
in order to discard and replace compromised templates. Furthermore,
those templates should not be related to one another, in the sense
that they should not be positively matched by the biometric system,
to prevent the impersonation of a subject with a stolen template.
Consequently, \textit{renewability} of biometric templates is also
desired. It should be noted that both cross-matching and
renewability can be addressed at the same time if full
\textit{unlinkability} between templates belonging to the same
subject is granted.

The relevance of these concerns and the efforts being directed to
solve them within the biometric community are highlighted by some
recent special issues in journals, such as the IEEE Signal
Processing Magazine Special Issue on Biometrics Security and Privacy
Protection \cite{SPM15PrivacyBio}, the development of international
standards on biometric information protection, such as the ISO/IEC
IS 24745 \cite{ISO-IEC-24745-2011}, specific tracks on biometric
security \cite{bowyer15BTAS,ICB2016} or privacy-enhancing
technologies \cite{SP2016,2016IFIP} at international conferences,
recent publications
\cite{nandakumar15BTPPerformance,rane14standardBPT}
and PhD Thesis
\cite{nagar12phdThesis,marta16phdThesis},
or the EU FP7 projects TURBINE on Trusted Revocable Biometrics
Identities\footnote{\url{http://www.turbine-project.eu/}} and PIDaaS
on Private Identification as a
Service\footnote{\url{http://www.pidaas.eu/}}.

\section{Conclusions}
\label{sec:conc}

The present article has presented a comprehensive survey of inverse
biometric methods. The experimental findings in most of the works
described in Sect.~\ref{sec:invBio} show that an attack on different
biometric systems based on a wide variety of characteristics using
such reconstructed samples would have a high chance of success. In
addition, depending on the inversion algorithm, not only
one but several synthetic samples, visually different from each
other, can be generated, all matching the stored reference template.
Success chances of impersonating a particular subject are hence
increased. All these facts proof the feasibility of recovering or
reconstructing synthetic biometric samples from the information
stored in unprotected reference templates. And thereby answer
positively some of the questions posed in the previous section.


Further research and investment in this field is fostered by the new European Union General Deata Protection Regulation \cite{EU-Regulation-DataPrivacy-160427}, which defines biometric data as \emph{sensitive personal data}. Within this regulation,
\emph{personal data} is defined as \lq\lq \emph{any information
relating to an identified or identifiable natural person (\lq\lq
data subject''); an identifiable natural person is one who can be
identified, directly or indirectly, in particular by reference to an
identifier such as a name, an identification number, location data,
an online identifier or to one or more factors specific to the
physical, physiological, genetic, mental, economic, cultural or
social identity of that natural person}''. This means that
\emph{processing} of biometric data is subject to right of
\emph{privacy preservation}, where the notion of \emph{processing}
means \lq\lq \emph{any operation or set of operations which is
performed on personal data or on sets of personal data, whether or
not by automated means, such as collection, recording, organisation,
structuring, storage, adaptation or alteration, retrieval,
consultation, use, disclosure by transmission, dissemination or
otherwise making available, alignment or combination, restriction,
erasure or destruction}''.

Those definitions imply that, in order to grant the subject's
privacy, biometric information should be carefully protected both in
its stored form (i.e., biometric templates or references) and any
time it is used for verification purposes. Therefore, with the main
goal of developing secure and privacy preserving biometric
technologies, new standardization efforts are being currently
directed to prevent such information leakages. In particular, the
ISO/IEC IS 24745 on biometric information protection
\cite{ISO-IEC-24745-2011} encourages the substitution of traditional
biometric systems with biometric template protection schemes (see
Sect.~\ref{sec:btp}).

\small

\section*{Acknowledgements}
This work was supported by the German Federal Ministry of Education
and Research (BMBF) and the Hessen State Ministry for
Higher Education, Research and the Arts (HMWK) within ATHENE - National 
Research Center for Applied Cybersecurity.


\bibliography{references}

\end{document}